\documentclass[runningheads]{llncs}
\usepackage{graphicx}
\usepackage{amsmath,amssymb} 
\usepackage{color}

\usepackage{multirow,array}
\usepackage[table]{xcolor}
\usepackage{subfigure}
\usepackage[noend]{algorithm2e} 
\usepackage{dsfont}
\usepackage{lipsum}
\usepackage{balance} 
\usepackage{multirow,booktabs}
\usepackage{enumitem}
\usepackage{cite}
\graphicspath{ {./images/} }
\usepackage[pagebackref=true,colorlinks]{hyperref}

\begin{document}
\pagestyle{headings}
\mainmatter

\def\ACCV20SubNumber{75}  

\title{Adversarial Semi-Supervised \\Multi-Domain Tracking} 
\titlerunning{Adversarial Semi-Supervised MDL}
%
\author{Kourosh Meshgi\inst{1} \and
Maryam Sadat Mirzaei\inst{1}} 
\authorrunning{K. Meshgi \& M.S. Mirzaei}
%
\institute{RIKEN Center for Advanced Intelligence Project (AIP), Tokyo, Japan
\email{\{kourosh.meshgi, maryam.mirzaei\}@riken.jp}
}

\maketitle

\begin{abstract}
Neural networks for multi-domain learning empowers an effective combination of information from different domains by sharing and co-learning the parameters. In visual tracking, the emerging features in shared layers of a multi-domain tracker, trained on various sequences, are crucial for tracking in unseen videos. Yet, in a fully shared architecture, some of the emerging features are useful only in a specific domain, reducing the generalization of the learned feature representation. We propose a semi-supervised learning scheme to separate domain-invariant and domain-specific features using adversarial learning, to encourage mutual exclusion between them, and to leverage self-supervised learning for enhancing the shared features using the unlabeled reservoir. By employing these features and training dedicated layers for each sequence, we build a tracker that performs exceptionally on different types of videos. 
\dots
\end{abstract}

\section{Introduction}
\label{sec:intro}
Multi-task learning (MTL) is a branch of supervised learning that strives to improve the generalization of the regression or classification task by leveraging the domain-specific information contained in the training signals of related tasks \cite{caruana1993multitask}. MTL has been investigated in various applications of machine learning, from natural language processing \cite{collobert2008unified} and speech recognition \cite{deng2013new} to computer vision \cite{girshick2015fast}. The tasks can be defined as applying the same model on different data (aka multi-domain learning, MDL) \cite{nam2016learning,liu2017adversarial}, or on various problems \cite{zamir2018taskonomy}. In NN-based MDL, different domains share a set of parameters, which means more training data for the shared feature space, faster training, and better generalization by averaging inherent noise of different domains \cite{ruder2017overview}.

\begin{figure}[!t]
\centering
\subfigure[\small FS MDL \label{fig:mtl1}]{\includegraphics[width=0.32\linewidth]{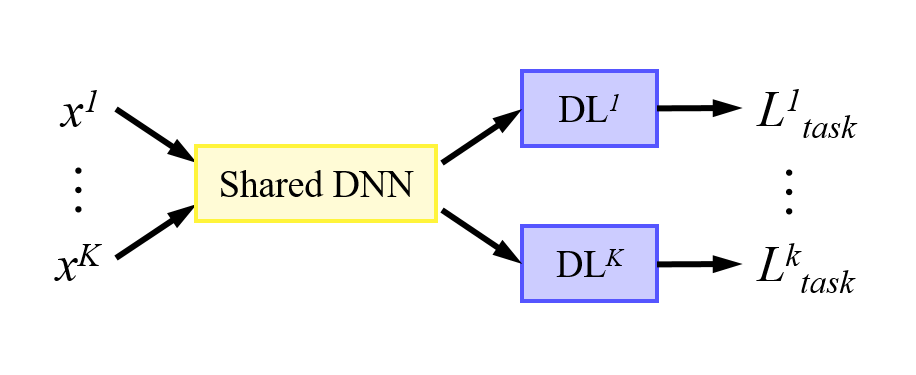}} 
\subfigure[\small PS MDL \label{fig:mtl2}]{\includegraphics[width=0.32\linewidth]{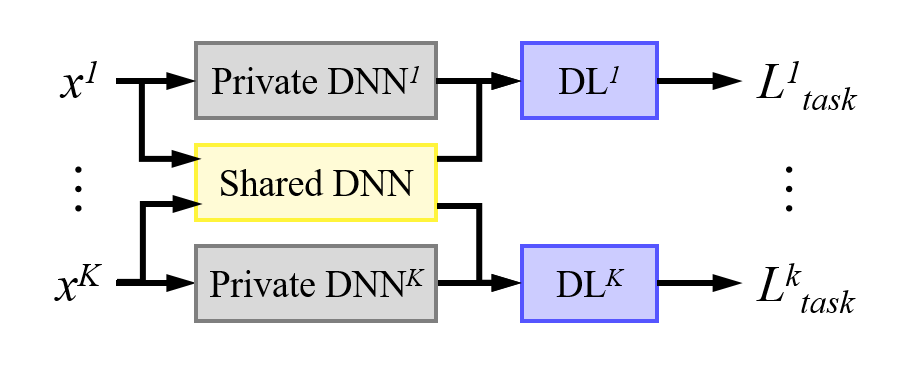}}
\subfigure[\small ASPS MDL \label{fig:mtl3}]{\includegraphics[width=0.33\linewidth]{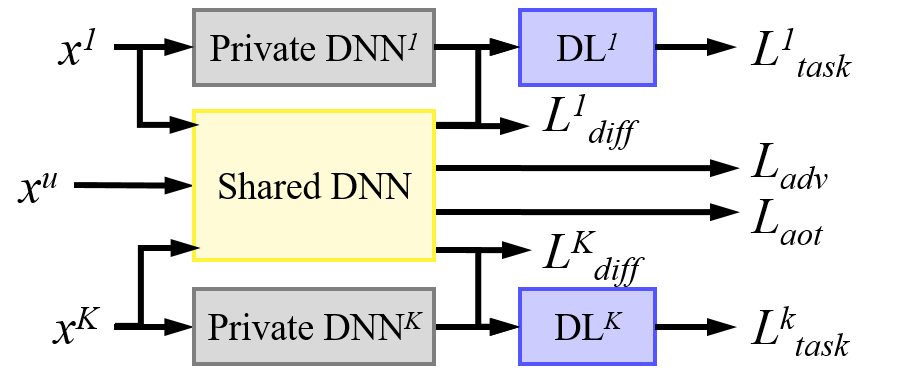}}
\caption{Different methods for obtaining shared representations using multi-domain learning (MDL). Panel (c) depicts our proposed architecture, Adversarial Semi-supervised Private-Shared (ASPS) MTL including adversarial, mutual exclusion, and self-supervised signals that promote the learning task.}
\label{fig:intro}
\vspace{-0.8 cm}
\end{figure}

MDNet \cite{nam2016learning} introduced this form of learning into a visual tracking problem, in which different video sequences are considered as different domains to learn from, and the task is defined as a foreground-background separation. 
This method, however, suffers from several setbacks: \textit{(i)} this model captures domain-independent representation from different domains with a fully-shared (FS) network (Figure \ref{fig:mtl1}). Such an architecture is unable to exclude the domain-specific features from the shared space, and either ignores them (underfit) or includes them in shared representation (overfit). In the former case, the training tries to compensate by non-optimally change the shared feature space. In contrast, in the latter, the learned feature in the shared space is merely useful for one or a few domains, wasting the model capacity for an inferior representation learning \cite{bengio2013representation}. Therefore, training on one domain inevitably hurts other domains and hinders the emergence of diverse shared features; \textit{(ii)} a shallow network is selected to avoid vanishing gradients, enabling training with limited number of annotated videos, assuming that tracking is relatively easier than object classification and requires less number of layers, and \textit{(iii)} the learned representation does not consider the patterns of the target motion. 

To address these issues, we proposed a private-shared (PS) MDL architecture that separates the domain-specific and domain-invariant (shared) features to disentangle the training on different domains and allow for learning an effective, shared feature representation (Figure \ref{fig:mtl2}). However, experiments show that PS architecture by itself is not enough to prevent domain-specific features from creeping into the share space \cite{ganin2015unsupervised}. Therefore, we proposed to have a virtual discriminator to predict which domain introduces the feature to the shared space. In a GAN-style optimization, we encourage the shared space to have only domain-independent features. We also introduced a regularization term to penalize redundancy in different feature spaces. To deal with the limited number of annotated video sequences, we employ self-supervised learning in which another virtual classifier using shared feature space is constructed to detect the playback direction (i.e., playing forward and backward) to enforce shared feature space to learn low-level features as well as semantics (Figure \ref{fig:mtl3}). The use of both supervised and unsupervised videos improves the discriminative power of shared feature space, as shown with the experiments. Further, we employed ST-ResNets \cite{feichtenhofer2016spatiotemporal} to learn spatiotemporal features, addressing the vanishing gradients, and separating domain-specific motion patterns from the shared ones. Finally, the learned features are used in a custom-made tracking-by-detection method to asses its transferability and effectiveness. 
In general we \textit{(i)} propose deep private-shared (PS) MTL architecture for tracking, 
\textit{(ii)} propose adversarial training for PS-MTL, 
\textit{(iii)} integrate semi-supervised and self-supervised training in a single MTL framework, 
\textit{(iv)} enforce learning spatiotemporal features in multi-domain tracking, using self-supervision and a backbone network capable of doing that, and
\textit{(v)} propose SUS-based hard negative batch mining, and
\textit{(vi)} conducted extensive ablation, design choice, and performance experiments.

The problem at hand is closely similar to \cite{sebag2019multi}, which tries to perform domain adaptation, leverage multiple datasets with overlapping but distinct class sets, and tries to separate labeled and non-labeled data using multi-domain adversarial learning. However, we explicitly divide features into shared and specific groups, to obtain features that generalize well and push out the specific features that reveal the originating domain in a GAN-style setup.\vspace{-0.5cm}

\section{Related Works}
\label{sec:literature}

\noindent\textbf{Deep Visual Tracking.} Early studies in the use of deep learning in tracking utilized features from autoencoders \cite{DLT,AE-ENS} and fully-connected layers of pre-trained (CNN-based) object detector \cite{fan2010human}. Still, later the CNN layers themselves were used to serve as features that balance the abstraction level needed to localize the target \cite{ma2015hierarchical}, to provide a spatiotemporal relationship between the target and its background \cite{CNT}, to combine spatial and temporal aspects of tracking \cite{UCT,YCNN}, and to generate probability maps for the tracker \cite{wang2015transferring}.
Recently the tracking studies pay attention to other deep learning approaches including the use of R-CNN for tracking-by-segmentation \cite{drayer2016object}, Siamese Networks for similarity measuring \cite{SINT,SIAMESEFC,li2018high,li2019siamrpn++} and GANs to augment positive samples \cite{wang2018sint++,VITAL}.

Most of the current CNN-based techniques use architectures with 2D convolutions to achieve different invariances to the variations of the images. Meanwhile, the invariance to transformations in time is of paramount importance for video analysis \cite{varol2018long}. Modeling temporal information in CNNs has been tackled by applying CNNs on optical flow images \cite{gkioxari2015finding}, reformulating R-CNNs to exploit the temporal context of the images \cite{chao2018rethinking} or by the use of separate information pathways for spatial and temporal pathways \cite{simonyan2014two,zhu2017end}. Motion-based CNNs typically outperform CNN representations learned from images for tasks dealing with dynamic target, e.g., action recognition \cite{varol2018long}. In these approaches, a CNN is applied on a 3-channel optical flow image \cite{dosovitskiy2015flownet}, and different layers of such network provide different variances toward speed and the direction of the target's motion \cite{feichtenhofer2018have}. In visual tracking, deep motion features provide promising results \cite{gladh2016deep}, which requires the tracker to fuse the information from two different CNN networks, a temporal and a spatial \cite{zhu2017end}. However, due to the difficulties in fusing these two streams, some researchers use only the motion features \cite{gladh2016deep}.

\noindent\textbf{Deep Multi-Task/Domain Learning.}
When used with deep learning, MTL models tend to share the learned parameters across different tasks either by \textit{(i)} hard parameter sharing \cite{caruana1993multitask} that the hidden layers are shared between all tasks, while task-specific ones are fine-tuned for each task, or by 
\textit{(ii)} soft parameter sharing, where each task has its parameters, and the distance between the parameters of the model is then regularized to encourage the parameters to be similar using, e.g., $\ell_1$ norm \cite{duong2015low} or trace norm \cite{yang2017trace}.

In the hard parameter sharing architectures, shared parameters provide a global feature representation, while task-specific layers further process these features or provide a complementary set of features suitable for a specific task. Some MTL approaches are based on the intuition that learning easy tasks is the prerequisite for learning more complex ones \cite{ruder2017overview}, hence put tasks in hierarchies \cite{sogaard2016deep,hashimoto2017joint,sanh2019hierarchical} or try to automatically group similar tasks to dynamically form shared layers \cite{liu2017distributed}. 
When training a multi-task learner, training each task normally increases the task's accuracy (fine-tuning) and, at the same time, provides more information for the shared representation that affects the accuracy of the rest of the tasks (generalization). Balancing the fine-tuning-generalization trade-off has been the subject of several studies. Kendall et al. \cite{kendall2018multi} adjusts tasks' relative weights in the loss function in proportion to the task uncertainty, \cite{liu2016recurrent} divides the feature space into task-specific and shared spaces and later employs adversarial learning to encourage shared feature space to contain more common and less task-specific information \cite{liu2017adversarial}, and \cite{bousmalis2016domain} proposed orthogonality constraints to punish redundancy between shared and task layers. In line with this, learning through hints \cite{abu1990learning} trains a network to predict the most important features. 
To leverage correlations of different problems in MTL, the tasks could be adversarial \cite{ganin2015unsupervised}, provide hints or attention for a main task \cite{yu2016learning,caruana1997multitask}, use a shared memory \cite{liu2016deep}, explicitly perform representation learning for a more complex task \cite{rei2017semi}, facilitate training for a quickly-plateauing main task \cite{bingel2017identifying}, or learn a shared unsupervised representation \cite{doersch2017multi}. It can be helpful to learn the relations between tasks to enable efficient transfer learning between them \cite{zamir2018taskonomy}.

\begin{figure}[!t]
\centering
\includegraphics[width=0.6\linewidth]{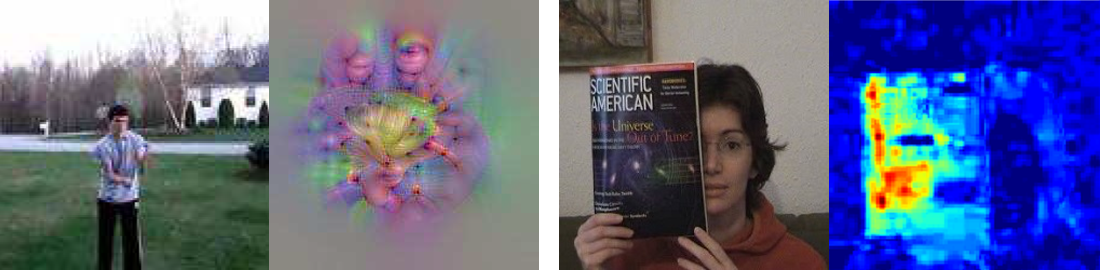}
\caption{Some of features in MDNet are activated for a few sequences containing a particular motion pattern or appearance and and wastes the capacity of shared feature space. Domain-specific features emerged in shared space (left) motion features maximally activated by motion pattern in \texttt{\small Joggling} (right) appearance feature maximally activated by \texttt{\small FaceOcc1} due to the book cover (visualized using \cite{feichtenhofer2018have} and \cite{zeiler2014visualizing}).
}
\label{fig:bad_examples}
\vspace{-0.5 cm}
\end{figure}

\noindent\textbf{Self-Supervised Learning.}
Self-supervised learning refers to a learning problem from unlabeled that is framed as a supervised learning problem by using proxy tasks on behalf of the main learning objective. Supervised learning algorithms are used to solve an alternate or pretext task, the result of which is a model or representation that can be used in the solution of the original (actual) modeling problem \cite{kolesnikov2019revisiting}, e.g., by masking a known word in a sentence and trying to predict it \cite{devlin2018bert}. Common examples of self-supervised learning in computer vision is to make colorful images grayscale and have a model to predict a color representation (colorization)\cite{zhang2016colorful}, removing blocks of the image and have a to model predict the missing parts (inpainting)\cite{doersch2015unsupervised}, or rotating an image by random multiples of 90$^{\circ}$ and predict if the image has the correct rotation \cite{gidaris2018unsupervised}. 
Self-supervised learning extracts additional information from video sequences by exploiting the natural structure in them to use as labels for proxy learning problems. Such information may derive from low-level physics (e.g., gravity), high-level events (e.g., shooting the ball in a soccer match), and artificial cues (e.g., camera motion) \cite{doersch2015unsupervised,wei2018learning}. The video sequence order \cite{misra2016shuffle}, video colorization \cite{vondrick2018tracking}, and video direction \cite{wei2018learning} have been explored as the pretext task to learn video representations. For this purpose, CNNs are more suitable to learn representations compared to autoencoders \cite{DEEPTRACK}. \vspace{-0.5cm}

\section{Proposed Tracker}
\label{sec:proposed}
This section describes the architecture of the proposed Adversarial Semi-supervised Multi-Domain (ASMD) tracker. We consider multi-domain (domain:=video) learning to share the acquired knowledge from one sequence to another. First, we explore the private-shared MDL framework that provides both domain-specific and shared features for the tracking-by-detection. The network architecture of private and shared modules to capture spatiotemporal features and their training is discussed next. To avoid redundancy and loss of generalization in the shared layers, we introduce adversarial regularization and orthogonality constraint into the MDL training (Fig. \ref{fig:schematic1}).
Further, to boost the performance of the MDL, the shared representation is intermittently trained using unlabeled videos (Fig. \ref{fig:schematic2}).

\subsection{Private-Shared Multi-Domain Learning}
In FS-MDL (Figure \ref{fig:mtl1}), all of the domains share the features extracted by a shared set of layers. Such an idea has been explored in MDNet tracker \cite{nam2016learning}, which assume that all features extracted for one domain is useful for other domains. This model ignores that some features are domain-dependent and may not be useful for other domains. Such features are maximally activated for one or very few domains (i.e., video sequences) while having low activation for others. Figure \ref{fig:bad_examples} illustrates some examples of such domain-specific features. 
On the other hand, PS-MDL \cite{liu2017adversarial} divides the feature space of each domain into a domain-invariant part that is shared among all domains, and a domain-specific part. As Figure \ref{fig:mtl2} depicts, each domain has its own set of features, that together with the shared features (i.e., concatenated with them), forms the domain features. 

\subsection{Network Architecture}
\begin{figure}[!t]
\centering
\includegraphics[width=0.75\linewidth]{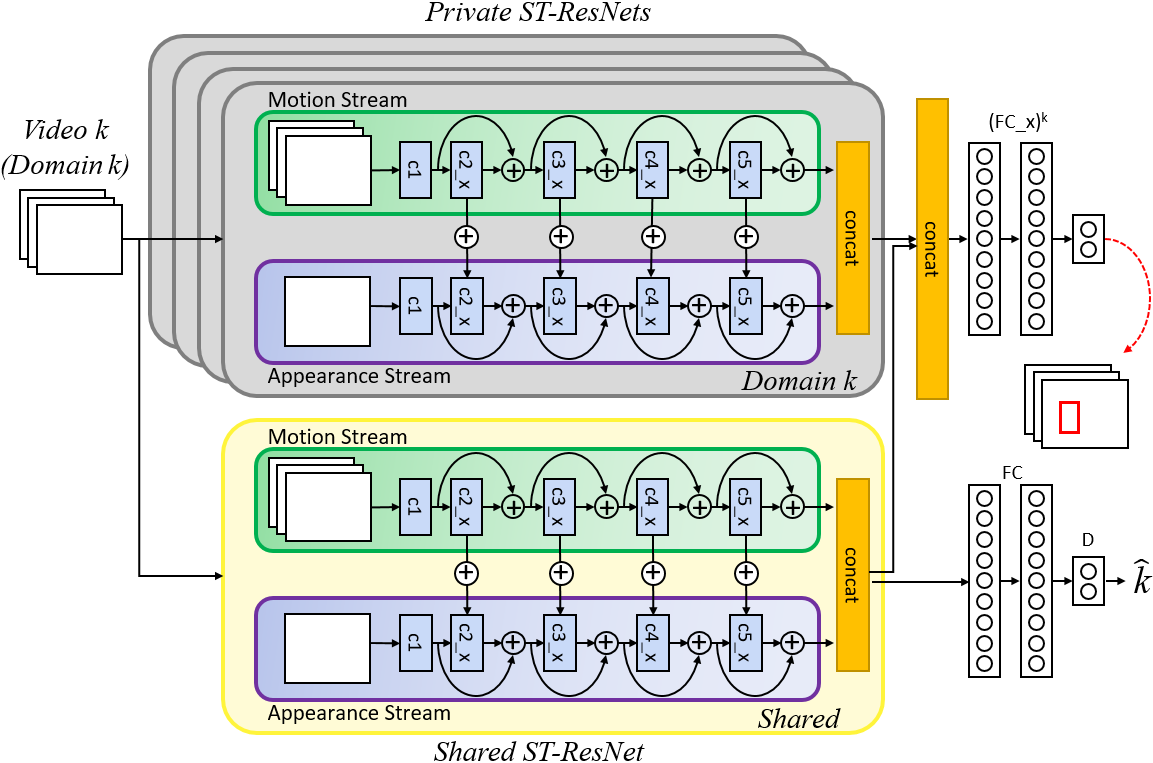}
\caption{The architecture of the proposed Private-Shared Multi-Domain Network, which consists of one shared and $K$ modified ST-ResNet domain-specific. For domain $k \in \{1..K\}$, the feature space consists of shared features and the corresponding domain-specific features, and the final fully connected layer calculated the classification score of the input patch from video $k$ for a foreground-background classification. }
\label{fig:schematic1}
\vspace{-0.5 cm}
\end{figure}

We used a modified Spatiotemporal ResNet \cite{feichtenhofer2016spatiotemporal} for our private and shared networks. This network uses a two-stream architecture \cite{simonyan2014two} each of streams having ResNet-18 architecture, and the final fully connected (FC) layer is dropped. The first layer of motion stream is altered to have $2L=10$ filter channels to operate on the horizontal and vertical optical flow stack of 5 frames. To establish the correspondence between the object and its motion, a residual link connects two streams for each ResNet block. The network receives 224$\times$224 input and reduces it to 7$\times$7 via global average pooling (GAP). Batch normalization and ReLU are applied after each convolution (Figure \ref{fig:schematic1}). The depth of the networks in MDNet \cite{nam2016learning} is kept low, however, with the use of ResNet blocks instead of ConvNet here, the network can be significantly deeper.

For each domain the shared features (i.e., concatenation of appearance and motion streams of shared layer) are concatenated with the domain-specific features, followed by two fully connected layers with 128 units and a binary classification fully-connected layer with softmax cross-entropy loss. The network is trained to minimize the cross-entropy of the predicted label and the true label of all target candidates for all domains. The loss is computed as:
\begin{equation}
\label{eq:loss_domain}
L_{dom}=\sum_{k=1}^{K} \alpha_{k} L_{track}\left(\hat{y}^{(k)}, y^{(k)}\right)
\end{equation}
where $\alpha_k$ is the weight of domain $k$ and $L_{track}\left(\hat{y}^{(k)}, y^{(k)}\right)$ is defined in eq\eqref{eq:loss_tracking}.

\subsection{Adversarial Representation Learning}
PS-MDL dissects the representation for each domain into a domain-specific and shared feature spaces but does not guarantee that shared feature space does not include domain-specific features and vice versa. To encourage the shared feature space to include more common information between domains yet has no domain-specific ones, we proposed to use adversarial training. Adversarial training became very popular after the seminal work of Goodfellow et al. \cite{goodfellow2014generative}. This approach sets up a min-max between networks in which a network $G$ tries to fool the other ($D$) by generating better samples (i.e., closer to real data distribution), and the other excels in distinguishing the generated data ($P_{G}$) from the real ones ($P_{data}$), by optimizing  
\begin{equation}
\ell =\min _{G} \max _{D}\left(\mathbb{E}_{x \sim P_{data}}[\log D(x)] +\mathbb{E}_{z \sim p(z)}[\log (1-D(G(z)))]\right)
\end{equation}

We propose an adversarial loss function for our MDL-based tracking. The goal is to encourage the shared space to be free from domain-specific knowledge. Therefore, we introduce a discriminator $D$ that predicts the domain $k$ only from the features that emerged in shared representation. The discriminator in each iteration is created from the learned shared ST-ResNet followed by three fully connected layers, each having 512 units and ReLU activation, and a $K$ unit softmax output layer. The multi-class adversarial loss \cite{liu2017adversarial} can be formulated as 
\begin{equation}
L_{adv}=\min _{\theta_{s}}\left(\max _{\theta_{D}}\bigg(\sum_{k=1}^{K} \sum_{i=1}^{N_{k}} \delta_{i}^{k} \log \left[D_{\theta_{D}}\left(\mathbb{E}(\mathbf{x}^{(k)})\right)\right]\bigg)\right)
\end{equation}
in which $\theta_D$ and $\theta_S$ are the parameters of the discriminator $D$ and shared layer respectively, $\delta_{i}^{k}$ indicates the domain of the current input $i$, and $\mathbb{E}(\mathbf{x})$ denotes shared representation of $\mathbf{x}$ followed by 2 fully connected layers and just before the discriminator layer. The intuition here is that given a sample from domain $k$, the shared feature space tries to alter its representation to better discriminate the sample, while the discriminator guesses which domain is the sample from. If the learned feature helps the discriminator to find the domain, it includes some domain-specific knowledge \cite{ben2010theory}, and the defined loss $L_{adv}$ punishes this representation.
It should be noted that $L_{adv}$ ignores the label $y^{(k)}$ of the input $\mathbf{x}^{(k)}$, allowing for unsupervised training of the shared layer, which is essential to realize the semi-supervised training of the network . 

A drawback of the PS-MDL is that the domain-specific and shared feature space may not be mutually exclusive, and redundant domain-invariant features may emerge in both \cite{jia2010factorized}. To alleviate this, inspired by \cite{bousmalis2016domain}, we compare the activations of the private and shared networks, and penalize the high co-activations of the corresponding neurons that are likely to encode a similar latent feature, 
\begin{equation}
L_{dif} = \sum_{k=1}^K \big\| (\mathbf{f}^s)^T \mathbf{f}^k \big\|^2_F
\end{equation}
in which $\mathbf{f}^s$ and $\mathbf{f}^k$ are shared and domain-specific feature vectors (i.e., last layer neuron activations) for a given input and $\|.\|^2_F$ denoted the Frobenius norm. \vspace{-0.25cm}

\subsection{Self-Supervised Representation Learning}
To enrich the share representation, we want to leverage the structure of the video data to use them as labels for closely related proxy problems that can be solved with the current shared architecture. We select the video direction (forward or backward) as a proxy learning task that is compatible with this architecture.

The order of the frames in a natural video contains spatiotemporal clues about the way objects move. The order of the frames can provide a trivial label for a proxy classification task: to classify if the movie as played in the natural direction or backward. During this task, the network learns spatiotemporal features, which is useful for other tasks such as video understanding, action recognition, and object tracking. Therefore, we select random clips from our dataset, randomly invert the order of some of them, and give them to our network to classify. This helps the features to emerge in the shared layers of our network.

\begin{figure}[!t]
\centering
\includegraphics[width=0.75\linewidth]{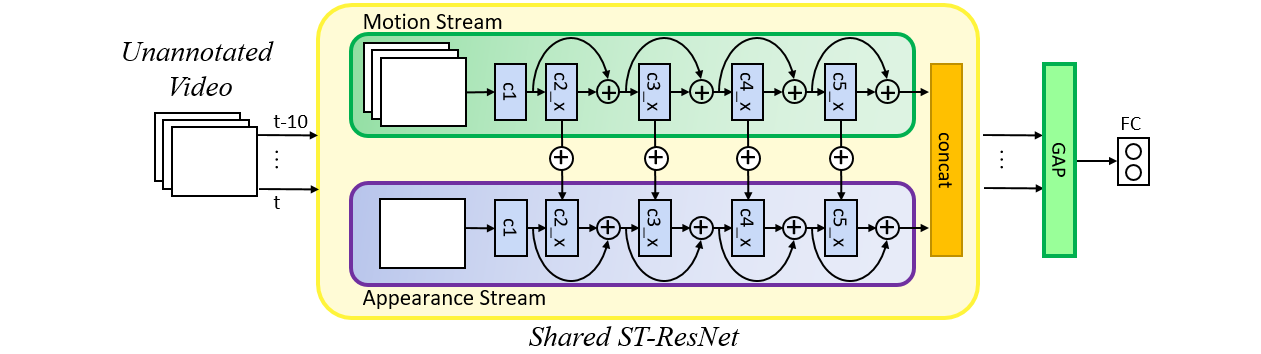}
\caption{Self-supervised network using shared ST-ResNet layers, a global average pooling layer, and an output layer to classify ``arrow of time'' in the 10-frame input clip.}
\label{fig:schematic2}
\vspace{-0.5 cm}
\end{figure}

To pull out this task, we draw 10-frame samples from annotated sequences as well as numerous unlabeled videos in YouTubeBB\cite{real2017youtube} dataset, randomly reverse their order (to augment the data with negative samples), and feed them to the shared ST-ResNet. The extracted features are then aggregated similar to \cite{wei2018learning} using a GAP layer, and a binary cross-entropy output layer computes the final arrow-of-time loss $L_{aot}$ for the forward-backward classification (Figure \ref{fig:schematic2}).

\subsection{Semi-Supervised MDL Training}
\label{sec:semi}
The final loss function of the tracker can be written as
\begin{equation}
L = L_{dom} + \lambda_1 L_{adv} + \lambda_2 L_{dif} + \lambda_3 L_{aot}
\end{equation}
in which $\lambda_1,\lambda_2,\lambda_3$ are hyper-parameters. This network is periodically trained on labeled and unlabeled videos such that after every 100 supervised iterations, 1000 unsupervised iterations are conducted. 

\noindent\textbf{Supervised.} For the supervised multi-domain learning, we collect 50 positive samples and 200 negative samples for each frame such that each positive sample has at least 70\% overlap with the target ($IoU \ge 0.7$), and negative samples have $IoU \le 0.5$). For multi-domain representation learning, the network is trained for 1M iterations with learning rate 0.0001 for ST-ResNets and 0.001 for FC layers on GOT10K dataset\cite{huang2019got} and its grayscale version. The networks are trained using backpropagation through gradient reversal layer \cite{ganin2015unsupervised} to assist the minimax optimization. For each iteration, five random 10-frame clip and another five with reversed frame-order are extracted for $L_{aot}$.

\noindent\textbf{Unsupervised.} We also conduct 10M iteration of unsupervised training on 240k videos of the YouTubeBB dataset. Again in each iteration, five random 10-frame clips and another five reversed one is extracted from the input video to provide the $L_{aot}$, and the learning rate for ST-ResNet is set to 0.00001.

\section{Online Tracking}
After learning the shared features, the private branches are discarded, and 3 new FC layers and a binary softmax output layer is used for every new sequence. During tracking (test time), no domain-specific network is used/trained. All of FC layers (after shared features) are randomly initialized and trained on-the-fly. 
The goal of private networks is to capture features that are not captured by the shared network (by trying to overfit the input sequence), introduce them to the feature pool, and receive the features that are pushed out of the shared pool.

\subsection{Tracking-by-Detection}
For each test sequence, the final FC layers convert the domain-specific and shared feature for all $n$ samples of each $T_k$ frame of video sequence $k$ into classification score $\hat{y} \in [0,1]$, trained by minimizing the cross-entropy of the sample's label ${y} \in \{0,1\}$ and the score, by applying SGD on the loss
\begin{equation}
\label{eq:loss_tracking}
L_{track}(\hat{y}^{(k)}, y^{(k)})=-\frac{1}{nT_k}\sum_{t=1}^{T_k} \sum_{j=1}^n y_{t}^{j(k)} \log \left(\hat{y}_{t}^{j(k)}\right)
\end{equation}
while the shared layers are kept frozen during tracking. This is important to avoid over-fitting to the video sequence \cite{nam2016learning} and for computational efficiency. After training on the first annotated frame of the sequence, $n$ samples are extracted from each frame $t$ following a Gaussian distribution around the last known target position. The samples are classified using the shared features and dedicated FC layers, and the sample with highest classification score is considered as the new target location. To find a tight bounding box around the target, the target candidates (i.e., score $>0.5$) are used in bounding box regression similar to \cite{girshick2014rich}.

To update the tracker, a dual-memory \cite{meshgi2018efficient} is employed in which the network is updated every $\Delta^s$ frames (short-memory) with all positive and selected negative samples between $t$ and $t-\Delta^s$. In the long run, the network is likely to forget the initial samples (especially those obtained from user), thus the network is updated every $\Delta^l$ frames ($\Delta^l\gg \Delta^s$) with the all previous target estimations (if score $> 0.5$) and the most uncertain negative sample (score $\rightarrow0.5$).  \vspace{-0.5cm}

\subsection{Stochastic Universal Negative Mining}
\label{sect:sus}
In visual tracking, negative examples (samples) comes from \textit{(i)} background, or \textit{(ii)} occluders, distractors and patches that has small overlap with the target. Examples from the former group are usually redundant and trivial for classification, while the latter ones are important to make a good foreground-background classifier but are sampled insufficiently. To identify such negative examples, hard negative mining \cite{sung1998example} explicitly predict examples' labels and selects misclassified examples (i.e., false positives). Hard minibatch mining \cite{nam2016learning} selects random subsets of negative examples and selects ones with the lowest classification score as the negative training examples without the need to the examples' labels. 
%
%
\begin{figure}[!t]
\centering
\includegraphics[width=0.9\linewidth]{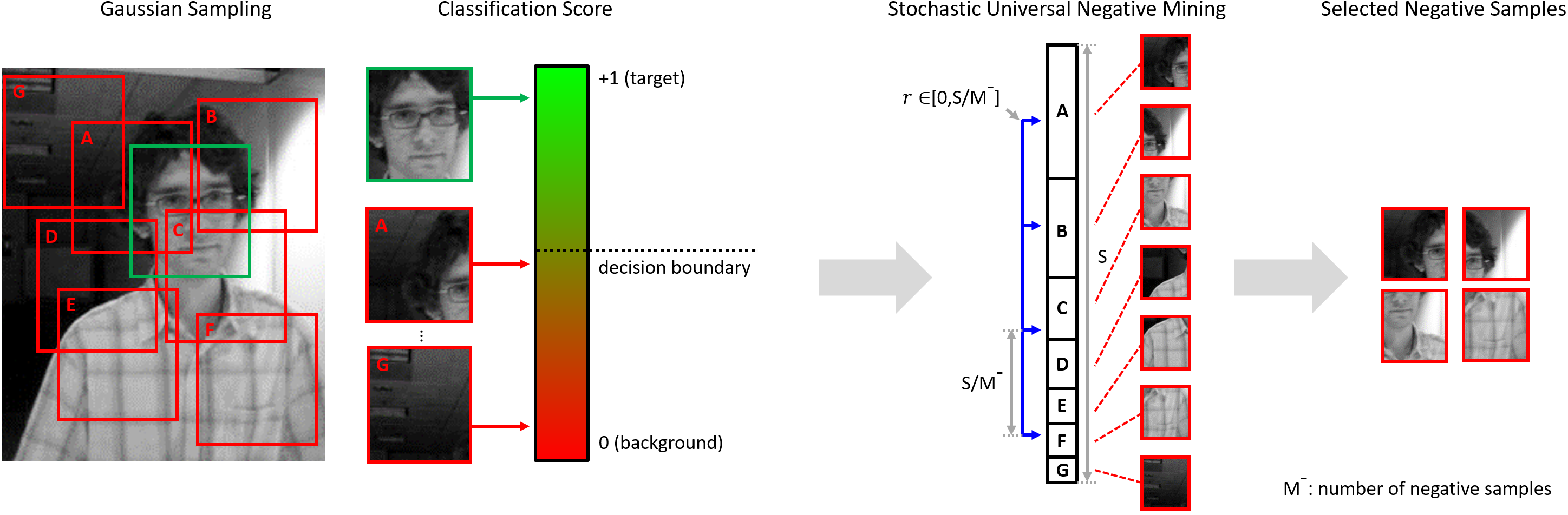}
\caption{Stochastic universal sampling to select negative examples for retraining. Using a comb-like ruler, SUS starts from a random real number, and chooses next candidates from the rest of negative examples, preventing the examples with the highest scores to saturate the selection space. $S$ is the sum of negative samples' classification scores.}
\label{fig:sus}
\vspace{-0.5 cm}
\end{figure}

To keep the focus on the hard negatives while keeping the diversity of the negative samples, we borrowed a stochastic universal sampling (SUS) technique \cite{baker1987reducing} from genetic algorithms to select negative training examples via repeated random sampling based on the classification score of the example (Figure \ref{fig:sus}). It should be noted that higher classification score for negative samples means that they are closer to the decision boundary (here, 0.5), thus the classifier learns more information about the sample distribution by training on them (uncertainty sampling principle \cite{lewis1994heterogeneous}). This technique gives weaker members of the negative example pool (according to their score) a chance to be chosen to represent different parts of the background in the classifier. \vspace{-0.5cm}

\subsection{Implementation Details}
This framework contains several sets of parameters. The tracking parameters including the number of positive and negative samples during train and test time, the covariance of the sampling from each frame, and the parameters of the bounding box regression are kept similar to MDNet\cite{nam2016learning}. The update intervals are set to $\Delta^s = 15$ and $\Delta^l = 50$. On the MDL side, all domains are equally weighted $\alpha_k=1/K$, and regularization hyper-parameters are tuned on GOT10K and YouTubeBB datasets via a grid search in $\lambda_i \in [0.01,0.1]$ and kept fixed through test-time trackings/experiments. Final values are {0.08,0.01,0.6} for $\lambda_{1..3}$.

The FC layers of tracker during the test are trained on the first (annotated) frame for 30 iterations with the learning rate of 0.0001 for hidden layers and 0.001 for the softmax layer. To update the classifier, the learning rate is increased to 0.0003 and 0.003 for hidden and output FC layers for ten iterations, with mini-batches composed of 32 positive and 96 SUS-sampled negatives out of 1024 available negative examples. The tracker is implemented in Matlab using MatConvNet \cite{vedaldi2015matconvnet} and ST-ResNet Github Repos, and runs around 5.18 fps on Intel Core i7 @ 4.00GHz with Nvidia V100 GPU. \vspace{-0.5 cm}

\section{Experiment}
\label{sec:eval}
To fully explore the capabilities of the proposed multi-domain learning for visual tracking, several experiments are conducted. In a series of experiments using OTB50 dataset \cite{wu2013online}, the effect of the MDL architecture and the network used in it, and the proposed negative minibatch mining are investigated. An ablation study is conducted to clarify the contributions of each component in the tracker. Finally, the performance of the tracker on recent challenging datasets such as OTB-100 \cite{wu2015object}, LaSOT \cite{fan2019lasot}, UAV123 \cite{mueller2016benchmark}, TrackingNet \cite{muller2018trackingnet}, and VOT-2018 \cite{kristan2018sixth} is benchmarked against the state-of-the-art trackers. For this comparison, we have used success and precision plots, where their area under curve provides a robust metric for comparing tracker performances \cite{wu2013online}. We also compare all the trackers by the success rate at the conventional thresholds of 0.50 ($IoU > \frac{1}{2}$) \cite{wu2013online}. For the state-of-the-art comparisons, the results are averaged for five different runs using the original semi-supervised training (\textit{cf} Section \ref{sec:semi}). The tracker is compared with latest trackers (DiMP \cite{bhat2019learning}, ATOM \cite{danelljan2019atom}, STResCF \cite{zhu2019stresnet_cf}, CBCW \cite{zhou2018efficient}), those with bounding box regression (SiamRPN\cite{li2018high}, SiamRPN++\cite{li2019siamrpn++}), those performing adversarial learning (VITAL \cite{VITAL}, SINT++ \cite{wang2018sint++}), and those using MTL (MDNet\cite{nam2016learning}, RT-MDNet\cite{jung2018realt}) as well as other popular trackers. 

\noindent\textbf{Multi-Domain Learning Model.}
Moving from FS-MDL in MDNet \cite{nam2016learning} to PS-MDL enables the model to separate the domain-invariant features from domain-specific ones. Adversarial and mutual-exclusion regularizations encourage this effect. The proposed adversarial training cannot operate on the fully-shared architecture of MDNet. Additionally, MDNet's backbone (VGG-M) is unable to learn spatiotemporal features, thus ineffective for our self-supervised learning.

Table \ref{tab:eval_mdl} presents five variations for the MDL architecture 
\textit{(i)} vanilla FS (different from MDNet), that uses only a fully shared ST-ResNet, 
\textit{(ii)} PS, that uses a task-specific ST-ResNet for each domain on top of the shared ST-ResNet, 
\textit{(iii)} PS+dif, that adds $L_{dif}$ to the PS variation, 
\textit{(iv)} APS-, that adds $L_{adv}$ to the PS variation, and 
\textit{(v)} APS that uses both $L_{adv}$ and $L_{dif}$ for the network. 
All variations are \textbf{only} trained with supervised data for 200k iterations, and regularization weights are tuned by 5-fold cross-validation for each case. 
Using PS-MDL without regularizations is not always beneficial, but using both in APS significantly improves the learning performance. MDNet (3 conv blocks) converges better than ST-ResNet (2$\times$5 conv blocks + skip connections) with similar iterations of training ($\sim10^5$) because of simpler architecture (significantly less parameters). While the performances of FS and PS are not as good as MDNet, when adding adversarial term (APS-) and the rest of regularizations (APS), the tracker outperforms MDNet (PS$<$FS$<$MDNet$<$APS-$<$APS).

\vspace{-0.5cm}
\begin{table}[!h]
\caption{Comparison of MDL models with 200k iterations of \textit{supervised} training on GOT10K \cite{huang2019got} and tested OTB-50. We also retrained MDNet on this dataset for the same number of iterations ($\star$). The {\color{red}first}, {\color{green}second} and {\color{blue}third} best methods are highlighted.}
\label{tab:eval_mdl}
\centering
\scalebox{0.7}{
\renewcommand{\arraystretch}{1.1}
\begin{tabular}{@{}l c c c c c c@{}}
\hline
 					& FS 	& MDNet$^\star$ 	& PS &	PS+dif & APS- 		& APS 	\\ \hline
Average Success  	& 0.62	& {\color{blue}0.67} &0.59	&0.41 &{\color{green}0.71}		& {\color{red}0.73}	\\
Average Precision	& 0.73	& {\color{blue}0.76} &0.70	&0.55 &{\color{green}0.80}		& {\color{red}0.81}	\\
$IoU>\frac{1}{2}$  	& 0.67	& {\color{blue}0.71} & 0.61	&0.45 &{\color{green}0.74}		& {\color{red}0.76}	\\
\hline
\end{tabular}
}
\vspace{-0.4 cm}
\end{table}

\noindent\textbf{Backbone Network.}
We tested our tracker with different backbone architecture such as MDNet-style VGG-M, Two Stream ConvNets with convolutional fusion \cite{feichtenhofer2016convolutional}, TwoStream ResNets with late fusion, Spatiotemporal ConvNet \cite{karpathy2014large} and SpatioTemporal ResNets\cite{feichtenhofer2016spatiotemporal}. Table \ref{tab:eval_backbone} shows that similar to the action recognition domain \cite{feichtenhofer2016spatiotemporal}, Spatiotemporal ResNets yields the best results for our intended tasks using to residual connections that enables learning the difference of sequences \cite{devlin2018bert}. Notice the big jump between single-stream and two-stream architectures that is partly because the arrow-of-time self-supervised training is mostly beneficial to train the motion stream.
\begin{table}[!t]
\caption{Comparison of backbone networks with 200k iterations of supervised training on OTB-50 (1S: single stream, 2S: two-stream, ST: spatiotemporal).}
\label{tab:eval_backbone}
\centering
\scalebox{0.65}{
\renewcommand{\arraystretch}{1.1}
\begin{tabular}{@{}l c c c c c @{}}
\hline
 					& 1S Conv~\cite{nam2016learning}~~~& 2S Conv~\cite{feichtenhofer2016convolutional}~~~& 2S Res~~~ & ST Conv~\cite{karpathy2014large}~~~ & ST Res~\cite{feichtenhofer2016spatiotemporal}~~~ 	\\ \hline
Avg Succ~~~  			& 0.64	& 0.69	& 0.67	& 0.63 	& {\color{red}0.73}	\\
Avg Prec			& 0.71	& 0.77	& 0.77	& 0.71	& {\color{red}0.81}	\\
$IoU>\frac{1}{2}$  	& 0.66	& 0.70	& 0.69	& 0.66	& {\color{red}0.76}	\\
\hline
\end{tabular}
}
\vspace{-0.5 cm}
\end{table}

\noindent\textbf{Negative Minibatch Selection.} To select representative negative samples from the pool, and to avoid model drift by insufficiently sampling from critical distractors and occluder, we proposed stochastic universal negative mining (SUNM). Table \ref{tab:eval_negative} shows the effectiveness of this method compared to a random selection of negative samples (RAND), hard minibatch mining (HMM) \cite{nam2016learning}, and a greedy selection of the most uncertain negative examples (GRDY).

\begin{table}[!t]
\caption{Comparison of negative minibatch selection for tracking on OTB-50.}
\label{tab:eval_negative}
\centering
\scalebox{0.65}{
\renewcommand{\arraystretch}{1.1}
\begin{tabular}{@{}l c c c c@{}}
\hline
 					& RAND~~ 	& GRDY~~ 	& HMM\cite{nam2016learning}~ 	& SUNM (\textbf{ours}) 	\\ \hline
Average Success  	& 0.72	& 	0.61	& 	0.77	& 	{\color{red}0.80}	\\
Average Precision	& 0.81	& 	0.69	& 	0.89	& 	{\color{red}0.88}	\\
$IoU>\frac{1}{2}$  	& 0.75	& 	0.67	& 	0.87	& 	{\color{red}0.93}	\\
\hline
\end{tabular}
}
\vspace{-0.3 cm}
\end{table}
\begin{table}[!t]
\caption{Comparison of the transferability of learned representations with different versions of the proposed framework as well as the pre-training on ILSRVC-2012 action recognition task for leave-one-out evaluation. For each video sequence of OTB50, 49 is used for semi-supervised training, and the remaining video is used for tracking, and the average of the success and precision of the tracker is reported in the table.}
\label{tab:eval_transfer}
\centering
\scalebox{0.65}{
\renewcommand{\arraystretch}{1.1}
\begin{tabular}{@{}l c c c c c@{}}
\hline
 					& PRE~~ 	& LOO-FS~~ 	& LOO-PS~~ 	& LOO-APS~~ 	& LOO-ASPS 	\\ \hline
Avg Succ  			& 0.41	& {\color{blue}0.70}		& 0.67	& {\color{green}0.72}	& {\color{red}0.78} \\
Avg Prec			& 0.52	& {\color{blue}0.76}		& 0.72	& {\color{green}0.80}	& {\color{red}0.85} \\
$IoU>\frac{1}{2}$  	& 0.46	& {\color{green}0.79} 		& 0.70	& {\color{green}0.79}	& {\color{red}0.87} \\
\hline
\end{tabular}
}
\vspace{-0.3 cm}
\end{table}
\noindent\textbf{Shared Knowledge Transfer.}
To test the transferability of the learned shared representation, we design a leave-on-out experiment on OTB50 dataset, in which out of 50 unique sequences in this dataset, we train the MDL on 49 of them for training, and the remaining one for testing the sequence using only 100k supervised and 10M unsupervised iterations, and the shared representation is kept frozen during tracking (LOO-ASPS). This process is repeated for three ablated versions of the proposed model, \textit{(i)} with fully-shared architecture (LOO-FS), \textit{(ii)} with the private-shared model but without the regularization terms (LOO-PS), and \textit{(iii)} without unsupervised iterations (LOO-APS), all trained with 100k supervised iterations. We also set the parameters of shared layers with the ILSRVC-2012 pre-trained network (PRE). Table \ref{tab:eval_transfer} shows that even FS-MDL in the LOO setting outperforms the tracker made with pre-training for the action recognition task. Also, PS-MTL without the regularization terms achieves inferior results compared to FS-MDL. PS-MDL with regularization but without unsupervised iterations outperform FS-MDL, however, when unsupervised training is used (hence, the name semi-supervised), such distinction becomes more apparent. In summary, it is shown that the full treatment yields better results due to multi-task learning (PRE$<$LOO-FS), adversarial training (LOO-PS$<$LOO-FS$<$LOO-SPS), and self-supervised training (LOO-APS$<$LOO-ASPS).

\noindent\textbf{Tracking Challenges Analysis.}
Table \ref{tab:attributes} presents the performance of the proposed tracker under challenging subsets of OTB50. The results reveal that using proposed PS-MTL architecture as well as the semi-supervised training scheme almost improved every aspect of the tracker compared to the baseline (MDNet). This improvement is more significant for occlusions, out-of-view and motion challenges due to the usage of spatiotemporal features in the tracker.
\begin{table}[!t]
\caption{Quantitative evaluation of trackers under different visual tracking challenges of OTB50 \cite{wu2013online} using AUC of success plot and their overall precision. 
}
\label{tab:attributes}
\centering
\scalebox{0.65}{
\renewcommand{\arraystretch}{1}
\begin{tabular}{@{}l ccccccccccc@{}}
\hline
{Attribute} & {\small TLD } & {\small STRK } & {\small TGPR } & {\small MEEM }& {\small MUSTer } & {\small STAPLE } & {\small CMT } &{\small SRDCF }& {\small CCOT }& {\small MDNet } & {\small Ours} \\ 
 & {\small\cite{kalal2012tracking}} & {\small\cite{hare2011struck}} & {\small \cite{gao2014transfer}} & {\small \cite{zhang2014meem}}& {\small \cite{hong2015multi}} & {\small  \cite{bertinetto2016staple}} & {\small  \cite{meshgi2017active}} &{\small \cite{danelljan2015learning}}& {\small \cite{danelljan2016beyond}}& {\small  \cite{nam2016learning}} &  \\ \hline
Illumination Variation     & 0.48 & 0.53 & 0.54 & 0.62 & 0.73 & 0.68 & 0.73 & 0.70 & {\color{blue}0.75} &{\color{green}0.76}& {\color{red}0.80} \\
Deformation    & 0.38 & 0.51 & 0.61 & 0.62 & 0.69 & {\color{blue}0.70} & 0.69 & 0.67 & 0.69 &{\color{green}0.73}& {\color{red}0.74} \\
Occlusion    & 0.46 & 0.50 & 0.51 & 0.61 & 0.69 & 0.69 & 0.69 & 0.70 & {\color{green}0.76} &{\color{blue}0.75}& {\color{red}0.81} \\
Scale Variation     & 0.49 & 0.51 & 0.50 & 0.58 & 0.71 & 0.68 & 0.72 & 0.71 & {\color{blue}0.76} &{\color{green}0.78}& {\color{red}0.82} \\
In-plane Rotation    & 0.50 & 0.54 & 0.56 & 0.58 & 0.69 & 0.69 & {\color{blue}0.74} & 0.70 & 0.72 &{\color{green}0.75}& {\color{red}0.78} \\
Out-of-plane Rotation    & 0.48 & 0.53 & 0.54 & 0.62 & 0.70 & 0.67 &  0.73 & 0.69 & {\color{blue}0.74} &{\color{green}0.76}& {\color{red}0.80} \\
Out-of-View (Shear)     & 0.54 & 0.52 & 0.44 & 0.68 & 0.73 & 0.62 & 0.71 & 0.66 & {\color{green}0.79} &{\color{green}0.79}& {\color{red}0.84} \\
Low Resolution     & 0.36 & 0.33 & 0.38 & 0.43 & 0.50 & 0.47 & 0.55 & 0.58 & {\color{blue}0.70} &{\color{red}0.72}& {\color{green}0.71} \\
Background Clutter     & 0.39 & 0.52 & 0.57 & 0.67 & {\color{blue}0.72} & 0.67 & 0.69 & 0.70 & 0.70 &{\color{green}0.76}& {\color{red}0.77} \\
Fast Motion     & 0.45 & 0.52 & 0.46 & 0.65 & 0.65 & 0.56 & 0.70 & 0.63 & {\color{blue}0.72} &{\color{green}0.73}& {\color{red}0.78} \\
Motion Blur     & 0.41 & 0.47 & 0.44 & 0.63 & 0.65 & 0.61 & 0.65 & 0.69 & {\color{green}0.72} &{\color{green}0.72}& {\color{red}0.78} \\
\hline
\hline
Avg. Succ    & 0.49 & 0.55 & 0.56 & 0.62 & 0.72 & 0.69 & 0.72 & 0.70 & {\color{blue}0.75} &{\color{green}0.76}& {\color{red}0.80} \\
Avg. Prec 	 & 0.60 & 0.66 & 0.68 & 0.74 & 0.82 & 0.76 & 0.83 & 0.78 & {\color{blue}0.84} &{\color{green}0.85}& {\color{red}0.88} \\ 
$IoU>0.5$    & 0.59 & 0.64 & 0.66 & 0.75 & 0.86 & 0.82 & 0.83 & 0.83 & {\color{blue}0.90} &{\color{red}0.93}& {\color{red}0.93} \\
\hline
\end{tabular}
}
\vspace{-0.5 cm}
\end{table}

\noindent\textbf{Evaluation on OTB-100.}
We benchmarked the tracker against the competing trackers on OTB100 \cite{wu2015object} as presented in Table \ref{tab:eval_otb100}. In comparison with trackers who perform BB regression, those using adversarial learning, and the state-of-the-art in the tracking, our proposed tracker shows beter performance.

\begin{table}[!t]
\caption{Quantitative evaluation on OTB100 using success rate and precision.}
\label{tab:eval_otb100}
\centering
\scalebox{0.56}{
\renewcommand{\arraystretch}{1.1}
\begin{tabular}{@{}l@{} c c c c c c c |cccccccc| c@{}}
\hline
 					& {\small dSRDCF}	& {\small CCOT}			& {\small BACF}	&  {\small dSTRCF}& {\small STResCF}& {\small CBCW}&{\small ECO}& {\small SiamRPN}& {\small SiamRPN++}& {\small SINT++}& {\small VITAL}& {\small MDNet}& {\small RT-MDNet}& {\small ATOM}&{\small DiMP}&{\small \textbf{Ours}} \\ 
& {\small \cite{danelljan2015convolutional}}	& {\small \cite{danelljan2016beyond}}			& {\small \cite{kiani2017learning}}	&  {\small \cite{li2018learning}}& {\small \cite{zhu2019stresnet_cf}}& {\small \cite{zhou2018efficient}}& \cite{danelljan2017eco}&{\small \cite{li2018high}}& {\small \cite{li2019siamrpn++}}& {\small \cite{wang2018sint++}}& {\small \cite{VITAL}}& {\small \cite{nam2016learning}}& {\small \cite{jung2018realt}}& {\small \cite{danelljan2019atom}}& \cite{bhat2019learning} \\ \hline
Avg. Succ$\uparrow$ & {\color{green}0.69}& {\color{black}0.68} 	& 0.62 			& 0.68 & 0.59 & 0.61 &{\color{green}0.69}& 0.63 & {\color{green}0.69} & 0.57 & {\color{black}0.68} & 0.67 & 0.65 & 0.66 &0.68& {\color{red}0.73} \\
Avg. Prec$\uparrow$ & 0.81 				& {\color{black}0.85} 	& 0.82    			& -    & 0.83 & 0.81 &{\color{red}0.91}& 0.85 & {\color{red}0.91} & 0.76 & {\color{red}0.91} & 0.90 & 0.88 & - &0.89&{\color{red}0.91} \\
$IoU>\frac{1}{2}$   & 0.78 				& {\color{red}0.88} 	& 0.77 			& 0.77 & 0.76 & 0.76 &-& 0.80 & {\color{black}0.83} & 0.78 & 0.75 & 0.80 & 0.79 & {\color{blue}0.86} &0.87&{\color{red}0.88} \\
\hline
\end{tabular}
}
\vspace{-0.5 cm}
\end{table}
\noindent\textbf{Evaluation on VOT-2018.}
Table \ref{tab:eval_vot18} shows the comparison of our method with the competing algorithms on 60 challenging videos of VOT-2018 \cite{kristan2018sixth}.
\begin{table}[!t]
\caption{Evaluation on VOT2018 by expected avg overlap, robustness and accuracy.}
\label{tab:eval_vot18}
\centering
\scalebox{0.65}{
\renewcommand{\arraystretch}{1.1}
\begin{tabular}{@{}lccccccc|cccc|c@{}}
\hline
& {\small STURCK} & {\small MEEM}& {\small STAPLE} & {\small SRDCF} & {\small CCOT} &{\small SiamFC}&  {\small ECO}& {\small SiamRPN}& {\small SiamRPN++}& {\small ATOM}& {\small DiMP}&{\small \textbf{Ours}} \\
& {\small \cite{hare2011struck}}	&  {\small \cite{zhang2014meem}}&{\small \cite{bertinetto2016staple}}	& {\small \cite{danelljan2015learning}}& {\small \cite{danelljan2016beyond}}& {\small \cite{SIAMESEFC}}&{\small \cite{danelljan2017eco}}& {\small \cite{li2018high}}& {\small \cite{li2019siamrpn++}}& {\small \cite{danelljan2019atom}}& \cite{bhat2019learning} &\\ \hline
EAO$\uparrow$   & 0.097 & 0.192 & 0.169 & 0.119 & {\color{black}0.267} & 0.188 & {\color{black}0.280} & {\color{black}0.383} & {\color{blue}0.414} & {\color{black}0.401} & {\color{red}0.440} &{\color{green}0.427} \\
Acc$\uparrow$   & 0.418 & 0.463 & {\color{black}0.530} & 0.490 & {\color{black}0.494} & {\color{black}0.503} & 0.484 &{\color{black}0.586} & {\color{red}0.600} & {\color{black}0.590} & {\color{blue}0.597} & {\color{red}0.604}    \\
Rob$\downarrow$ & 1.297 & 0.534 & 0.688 & 0.974 & {\color{black}0.318} & 0.585 & {\color{black}0.276} &{\color{black}0.276} & {\color{black}0.234} & {\color{blue}0.204} & {\color{red}0.153} & {\color{green}0.169} \\   
\hline
\end{tabular}
}
\vspace{-0.5 cm}
\end{table}

\noindent\textbf{Evaluation on LaSOT.}
Evaluating our tracker on the LaSOT dataset \cite{fan2019lasot} has two phases. In phase I, our tracker is trained on the GOT10K dataset for 1M supervised iterations and on YouTubeBB \cite{real2017youtube} for 10M unsupervised iterations. As showed in table \ref{tab:eval_lasot}, our proposed algorithm obtained the best results compared to state-of-the-art trackers when tested on all 1400 video sequences of LaSOT. Protocol II brings an even more interesting challenge, that limits the training data to the given 1120 videos, and test the trained trackers on the remaining 280 sequences of the dataset. Using the given training videos, we conduct 1M supervised and 10M unsupervised training. The obtained shared layer obtained a significant improvement compared to the MDNet and the state-of-the-art, which can be attributed to the high amount of labeled data, and the close distributions of training and test data, which is essential for unsupervised training.

\begin{table}[!t]
\caption{Evaluation on LaSOT with protocol I (testing on all videos) and protocol II (training on given videos and testing on the rest). We get better results with dataset's own videos as training due to large training set and matching domain.}
\label{tab:eval_lasot}
\centering
\scalebox{0.65}{
\renewcommand{\arraystretch}{1.1}
\begin{tabular}{@{}l cccccc|ccccc|c@{}}
\hline
& {\small STAPLE} & {\small SRDCF} & {\small SiamFC}& {\small SINT} & {\small ECO}& {\small BACF}&  {\small SiamRPN++}& {\small VITAL}& {\small MDNet}&  {\small ATOM}&{\small DiMP}&{\small \textbf{Ours}} \\ 
& {\small \cite{bertinetto2016staple}}	& {\small \cite{danelljan2015learning}}			& {\small \cite{SIAMESEFC}}	&  {\small \cite{SINT}}& {\small \cite{danelljan2017eco}}& {\small \cite{kiani2017learning}}& {\small \cite{li2019siamrpn++}}& {\small \cite{VITAL}}& {\small \cite{nam2016learning}}& {\small \cite{danelljan2019atom}}& \cite{bhat2019learning} & \\ \hline
(I) Acc$\uparrow$     & 0.266 & 0.271 & 0.358 & 0.339 & 0.340 & 0.277 &{\color{black}0.496}&{\color{black}0.412}&{\color{black}0.413}&{\color{blue}0.515}& {\color{red}0.569} & {\color{green}0.554} \\
(I) Rob$\uparrow$  & 0.231 & 0.227 & {\color{black}0.341} & 0.229 & 0.298 & 0.239 &-&{\color{blue}0.372}&{\color{green}0.374}&-&-& {\color{red}0.487} \\
\hline
(II)Acc$\uparrow$    & 0.243 & 0.245 & 0.336 & 0.314 & 0.324 & 0.259 &-&{\color{blue}0.390}&{\color{green}0.397}&-&-& {\color{red}0.499}  \\
(II)Rob$\uparrow$  & 0.239 & 0.219 & {\color{black}0.339} & 0.295 & 0.301 & 0.239 &-&{\color{blue}0.360}&{\color{green}0.373}&-&-& {\color{red}0.495} \\ 
\hline
\end{tabular}
}
\vspace{-0.5 cm}
\end{table}

\noindent\textbf{Evaluation on UAV123.}
To measure the effectiveness of the tracker on the slightly different task, we conduct an experiment to track objects from low-altitude UAVs. Such videos are inherently different from traditional datasets like OTB100 and GOT10K, since they contain aerial viewpoint, relatively smaller targets, tilted horizon, more chance of the cluttered background, and higher degree of perspective effect. The result in table \ref{tab:eval_uav123} suggest that the shared feature space is good enough to surpass the state-of-the-art trackers for tracking in these videos, but not good enough to obtain a high success score. In this regard, we conducted an additional experiment by 5-fold cross validating the network trained for 100K supervised iterations and 1M unsupervised iteration for each fold (we ensured that each fold has exactly 4 of the 20 longer videos of the dataset, called UAV20L subset). The average success score for this task on UAV123 reaches 0.655 that is almost 7.8\% improvement compared to the generally trained version of the proposed tracker (on GOT10K and YouTubeBB). 
\begin{table}[!t]
\caption{Evaluation on UAV123 by success rate and precision.}
\label{tab:eval_uav123}
\centering
\scalebox{0.65}{
\renewcommand{\arraystretch}{1.1}
\begin{tabular}{@{}lccccccc|cccccc|c@{}}
\hline
& {\small STRUCK}& {\small MEEM} & {\small STAPLE} & {\small SRDCF} & {\small MUSTer} & {\small ECO} & {\small CCOT} & {\small SiamRPN}& {\small SiamRPN++}& {\small MDNet}& {\small RT-MDNet}& {\small ATOM}& {\small DiMP}&{\small \textbf{Ours}} \\
& {\small \cite{hare2011struck}}	&  {\small \cite{zhang2014meem}}&{\small \cite{bertinetto2016staple}}	& {\small \cite{danelljan2015learning}}& {\small \cite{hong2015multi}}&{\small \cite{danelljan2017eco}}& {\small \cite{danelljan2016beyond}}& {\small \cite{li2018high}}& {\small \cite{li2019siamrpn++}}& {\small \cite{nam2016learning}}& {\small \cite{jung2018realt}}& {\small \cite{danelljan2019atom}}& \cite{bhat2019learning}& \\ \hline
Succ$\uparrow$   &0.387 & 0.398 & 0.453 & 0.473 & 0.517 & {\color{black}0.522} & 0.513 & 0.527 & 0.613 & 0.528 & 0.528 & {\color{blue}0.643} & {\color{green}0.654} & {\color{red}0.655}\\
Prec$\uparrow$ & 0.578 & {\color{black}0.627} & 0.666     & {\color{black}0.676} & 0.391     & 0.591 &  - &{\color{black}0.748} & {\color{red}0.807} & -     & {\color{blue}0.772} & 0.856 & 0.858     & {\color{green}0.793}\\
\hline
\end{tabular}
}
\vspace{-0.25 cm}
\end{table}
\begin{table}[!t]
\caption{Evaluation on TrackingNet by precision, normalized precision, and success.}
\label{tab:eval_trackingnet}
\centering
\scalebox{0.65}{
\renewcommand{\arraystretch}{1.1}
\begin{tabular}{lccccccccccc}
\toprule
 & ECO & SiamFC & CFNet & MDNet & DaSiam-RPN & ATOM & SiamRPN++ & DiMP & \textbf{Ours}  \\
 & \cite{danelljan2017eco} & \cite{SIAMESEFC} &   & \cite{nam2016learning} & \cite{zhu2018distractor} & {\small \cite{danelljan2019atom}} & \cite{li2019siamrpn++} & \cite{bhat2019learning} &   \\
\midrule
Precision & 0.492 & 0.533 & 0.533 & 0.565 & 0.591 & {\color{blue}0.648} & {\color{red}0.694} & {\color{green}0.687} & {\color{green}0.687}  \\ 
Norm. Prec. & 0.618 & 0.666 & 0.654 & 0.705 & 0.733 & 0.771 & {\color{blue}0.800} & {\color{green}0.801} & {\color{red}0.802}  \\ 
Success (AUC) & 0.554 & 0.571 & 0.578 & 0.606 & 0.638 & 0.703 & {\color{blue}0.733} & {\color{green}0.740} & {\color{red}0.741}  \\
\bottomrule
\end{tabular}
}
\vspace{-0.5 cm}
\end{table}
\noindent\textbf{Evaluation on TrackingNet.} Table \ref{tab:eval_trackingnet} shows our results on the TrackingNet test set \cite{muller2018trackingnet} (511 videosh) that is comparable to the SOTA approaches.

\noindent\textbf{Discussion.}
Results shows that 
\textit{(i)} the proposed method outperforms VITAL and SINT++ that are most popular instances of using adversarial training in tracking to generate hard positive examples (unlike our method that used such training for obtaining better internal representation); 
\textit{(ii)} our method also outperforms different versions of SiamRPN. The reason is that the proposed method focused on learning more generalizable features while using a simple pre-trained RPN whereas SiamRPN trains the RPN networks with regards to extracted features with pre-trained conv layers. In this view, two methods (ours and SiamRPN++) can be combined to create a tracker with better features and better tracking method; and 
\textit{(iii)} our method outperformed MDNet using its adversarial and ST semi-supervised training. This table shows a significant improvement in OTB100, best results on UAV123, and competitive results on VOT2018. The performance drop in VOT2018 is due to frequent target disappearances that challenges our pre-trained RPN network.   \vspace{-0.5 cm}

\section{Conclusion}
\label{sec:conclusion}
In this study, we proposed a semi-supervised private-shared MDL to learn the domain-invariant and domain-specific features from annotated videos, push the domain-specific features out of the shared feature space using an adversarial minimax regularization to avoid domain overfitting, reduce the overlap between private and shared features, and use the direction of video playing as a proxy learning task to further train the shared representation using unannotated videos. We then proposed a tracker that classifies the target from the background for each video that uses the learned feature representation, employ a dual-memory update scheme to train the dedicated fully-connected layers after the feature layer, and uses a proposed novel negative minibatch mining to increase the tracker's accuracy and keeping it vigilant for potential distractors and occluders.

We have conducted several experiments on the backbone network, MDL architecture, negative minibatch selection, the transferability of the learned features, and the performance of the proposed tracker in different tracking situations. The results of the benchmarks on different datasets demonstrate the superior performance of the proposed tracker compared to the state-of-the-art. Our next step is to tailor our adversarial learning and global search for long-term tracking as proposed in \cite{wang2019learning,huang2020globaltrack}.


\bibliographystyle{splncs}
\bibliography{refs}

\end{document}